# Joint Graph Convolution and Sequential Modeling for Scalable Network Traffic Estimation

Nan Jiang
Carnegie Mellon University
Pittsburgh, USA

Wenxuan Zhu
University of Southern California
Los Angeles, USA

Xu Han
Brown University
Providence, USA

Weiqiang Huang
Northeastern University
Boston, USA

Yumeng Sun*
Rochester Institute of Technology
Rochester, USA

*Abstract-This study focuses on the challenge of predicting network traffic within complex topological environments. It introduces a spatiotemporal modeling approach that integrates Graph Convolutional Networks (GCN) with Gated Recurrent Units (GRU). The GCN component captures spatial dependencies among network nodes, while the GRU component models the temporal evolution of traffic data. This combination allows for precise forecasting of future traffic patterns. The effectiveness of the proposed model is validated through comprehensive experiments on the real-world Abilene network traffic dataset. The model is benchmarked against several popular deep learning methods. Furthermore, a set of ablation experiments is conducted to examine the influence of various components on performance, including changes in the number of graph convolution layers, different temporal modeling strategies, and methods for constructing the adjacency matrix. Results indicate that the proposed approach achieves superior performance across multiple metrics, demonstrating robust stability and strong generalization capabilities in complex network traffic forecasting scenarios.*

## I. INTRODUCTION

With the rapid development of the internet and the widespread adoption of emerging technologies such as 5G, IoT, and cloud computing, global network infrastructures are facing unprecedented complexity and heavy load challenges. Network traffic, as a critical indicator of network operation and user behavior, holds significant value in areas such as operation and maintenance, resource scheduling, and network security [1]. Especially in the context of surging services and increasingly complex network structures, accurate and efficient traffic prediction has become a key technology to ensure stable network operation and improve service quality. However, traditional traffic prediction methods are mostly based on time-series modeling. They struggle to capture topological correlations and spatial interactions among multiple nodes, leading to unsatisfactory performance in real-world scenarios. Therefore, it is imperative to develop prediction models that integrate network structure information to better characterize traffic patterns [2].

In recent years, deep learning has achieved significant advancements in handling graph-structured data. Specifically, Graph Convolutional Networks (GCNs) have shown excellent performance in extracting features from non-Euclidean domains [3]. Unlike conventional neural networks, GCNs can capture spatial dependencies among nodes within a network graph, thereby improving the ability to perceive and model node state evolution [4]. In network traffic prediction, GCNs leverage network topology to effectively capture inter-node interactions, offering a new approach for accurate traffic forecasting in multi-source heterogeneous networks. However, most existing GCN models are designed for relatively regular or simplified network structures [5]. They are not well-adapted to the dynamic topology and strong heterogeneity of real-world networks, which limits their applicability in practical scenarios.

In real-world settings, network topologies often exhibit high complexity, multi-level structures, and dynamic evolution. For example, in edge computing and multi-access environments, device nodes are highly heterogeneous, with unstable connections and diverse communication paths. These characteristics impose higher demands on traffic prediction models. Traditional GCN-based modeling approaches often overlook the impact of topological asymmetry, temporary link relations, and non-static adjacency matrices [6]. As a result, they lack the ability to effectively capture both global and local features. Therefore, building GCN-based traffic prediction algorithms tailored to complex topologies is not only a breakthrough in the adaptability and expressiveness of current models but also a technical foundation for addressing the challenges of dynamic network environments [7].

From an application perspective, improving traffic prediction accuracy in complex networks is of great practical significance. For example, in congestion warning systems, accurately forecasting traffic surges at hotspot nodes supports traffic scheduling and load balancing. In network security and anomaly detection, early prediction of abnormal traffic facilitates efficient threat detection and response [8-10]. In intelligent operation and maintenance systems, traffic prediction enables automated resource allocation and service quality optimization. Thus, designing spatially aware GCN models for complex topologies not only promotes intelligent network evolution but also provides a theoretical and algorithmic foundation for building robust, adaptive, and efficient network systems.

In summary, as network structures become increasingly complex and business traffic continues to evolve dynamically, traditional traffic prediction methods face significant limitations. GCNs offer a promising direction for network traffic modeling, yet their capabilities in handling complex topologies remain insufficient. This study aims to develop a GCN-based traffic prediction algorithm for complex network topologies by introducing multi-dimensional topology modeling mechanisms and dynamic structure-aware methods. This research not only holds significant theoretical value but also offers practical solutions for network traffic management and intelligent network services.

## II. METHOD

To effectively capture the complex spatial and temporal dynamics inherent in network traffic data, this study constructs a hybrid modeling framework that integrates Graph Convolutional Networks (GCNs) with a temporal processing module. The GCN component is employed to learn spatial dependencies between network nodes by operating on the graph structure of the network. This design is informed by the work of Y. Deng [11], who demonstrated the advantages of graph-based learning in handling complex data center topologies through reinforcement learning. For modeling the temporal dimension, the framework integrates a time-series module, enabling sequential learning of traffic evolution patterns over time. This design builds upon the insights of X. Wang [12], whose research into dynamic scheduling emphasized the importance of capturing temporal trends for resource optimization in computing environments. The overall model architecture, as depicted in Figure 1, follows a modular design to allow flexible integration and optimization of spatial and temporal components. This design also reflects principles from distributed computing environments, where task scheduling and communication efficiency are paramount to performance, particularly under federated learning settings [13], Together, these components form a joint architecture capable of robust traffic forecasting in large-scale and topologically diverse networks:

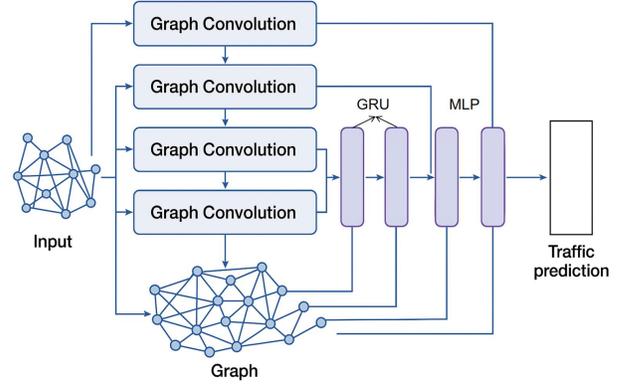

Figure 1. Overall model architecture

Figure 1 depicts the schematic flow of the proposed spatiotemporal learning model, consisting of three core components: Graph Convolutional layers for extracting spatial dependencies, GRU units for capturing temporal dynamics, and a Fully Connected Layer for generating predictive outputs. The modeling pipeline begins by encoding the network topology as a graph, which is processed through successive Graph Convolution layers to learn localized structural patterns among nodes. This spatial extraction phase is essential for capturing the static yet complex inter-node relationships typically found in distributed traffic systems, as emphasized in trust-aware scheduling frameworks for large-scale networks [14]. Subsequent to spatial encoding, the output feature sequences are input to a GRU-based temporal modeling module. The GRUs are tasked with learning the evolving nature of traffic patterns across time, an approach consistent with advanced sequence learning architectures that employ recurrent mechanisms for handling temporal variation in input data [15]. These techniques have proven effective in modeling sequential dependencies within fluctuating environments, especially when attention mechanisms or bidirectional encoders are integrated to enhance context sensitivity. The final stage involves a Multi-Layer Perceptron that translates the spatiotemporally processed features into traffic forecasts for future intervals. The use of an MLP here ensures that non-linear relationships between learned features and prediction targets are effectively captured, following principles found in high-dimensional pattern recognition systems [16]. This figure serves not only as a companion to the methodology section but also as a concise blueprint of how spatial structure and temporal behavior are fused within a single predictive framework tailored for complex network environments.

Assume that the entire network topology is represented as a graph $G = (V, E)$, where V represents the node set, E represents the edge set, and the topological structure can be represented by the adjacency matrix $A \in R^{N \times N}$, where N is the number of nodes. The feature of the node at time step t is represented as $X^{(t)} \in R^{N \times F}$, where F is the feature dimension of each node. In order to extract the spatial dependency in the graph structure, the graph convolution operation is used to

encode the input features. The basic graph convolution operation can be defined as:

$$H^{(l+1)} = \sigma(D^{-1/2} A' D^{-1/2} H^{(l)} W^{(l)})$$

Where $A' = A + I$ is the adjacency matrix with self-loops, $D$ is its corresponding degree matrix, $H^{(l)}$ represents the node representation of the l-th layer, $W^{(l)}$ is the trainable weight parameter, and $\sigma(\cdot)$ is the nonlinear activation function. Building on this foundation, to improve the model's adaptability to dynamically evolving topological structures, a Graph Attention Mechanism is incorporated to assign distinct weights to different neighboring nodes. The attention coefficient is defined as:

$$a_{ij} = \frac{\exp(\text{RELU}(a^T[wx_i \| Wx_j]))}{\sum_{k \in N(i)} \exp(\text{RELU}(a^T[Wx_i \| Wx_k]))}$$

Where $\|$ represents the concatenation operation, a and W are learnable parameters, and $N(i)$ represents the neighbor set of node i. This mechanism can adaptively focus on important adjacency relationships, thereby improving the expressiveness of graph representation.

Considering that network traffic has obvious temporal correlation, it is difficult to capture the evolution law in the time dimension using only the graph structure. Therefore, after extracting the spatial features, the gated recurrent unit (GRU) is introduced to model the node temporal features [17]. Let the node sequence obtained by graph convolutional encoding be $\{H^{(1)}, H^{(2)}, ..., H^{(T)}\}$, then the update process of GRU can be expressed as:

$$z_t = \sigma(W_z H^{(t)} + U_z h_{t-1})$$
$$r_t = \sigma(W_r H^{(t)} + U_r h_{t-1})$$
$$h'_t = \tanh(W_h H^{(t)} + U_h(r_t \otimes h_{t-1}))$$
$$h_t = (1 - z_t) \otimes h_{t-1} + z_t \otimes h'_t$$

$z_t$ and $r_t$ are the update gate and reset gate respectively, $\otimes$ represents the Hadamard product, and $h_t$ is the current hidden state. In the last layer, the output predicts the future node traffic value $Y'^{(T+1)}$ through the fully connected layer. In order to optimize the model, the mean square error loss function is used:

$$L = \frac{1}{N} \sum_{i=1}^{N} \| Y_i^{(T+1)} - Y'^{(T+1)}_i \|^2$$

By jointly optimizing graph spatial features and temporal evolution laws, the method proposed in this paper can effectively adapt to network traffic prediction tasks under complex topological structures and improve the overall prediction accuracy and generalization ability.

## III. EXPERIMENT

### A. Datasets

This study uses the widely adopted real-world network traffic dataset, the Abilene Dataset, for modeling and experimental validation. Abilene is a wide-area backbone network dataset provided by the Internet2 project. It includes 11 core routing nodes and a complete link topology. The dataset records traffic information for each link with temporal continuity and realistic network characteristics. It is commonly used in research on network traffic modeling and prediction.

The raw data is collected at 5-minute intervals, covering traffic dynamics over several days. Each time step contains bidirectional traffic values between multiple links, measured in bytes per second. For modeling and processing purposes, the data is normalized. A sliding window approach is applied to construct input and output sequences, enabling multi-step traffic prediction.

The Abilene dataset has a clear structure and well-defined topology. It reflects the spatiotemporal coupling of real network environments while maintaining sufficient stability and representativeness. It effectively supports graph-based modeling and time-series forecasting tasks. As a benchmark platform, it allows comprehensive evaluation of the proposed model's performance under complex network topologies.

### B. Experimental Results

This paper first conducted a comparative experiment on the prediction performance of different models, and the experimental results are shown in Table 1.

Table 1. Experiment on the impact of different number of topics on model performance

| Model | MAE | RMSE | R2 |
|---|---|---|---|
| Temporal GCN[18] | 2.86 | 5.42 | 0.928 |
| DCRNN[19] | 2.47 | 4.96 | 0.937 |
| AGCRN | 2.39 | 4.81 | 0.941 |
| MTGNN[20] | 2.26 | 4.53 | 0.948 |
| Ours(GCN+GRU) | 2.01 | 4.12 | 0.956 |

From the experimental results shown in the table, it can be observed that the performance of various deep neural network models improves to different extents as the model architectures evolve. Among them, the traditional Temporal GCN model shows relatively poor performance in terms of MAE and RMSE. This indicates that it has certain limitations in capturing temporal dependencies and spatial feature interactions.

In contrast, DCRNN achieves significantly lower errors by introducing a temporal recurrent mechanism. This demonstrates that enhancing temporal dynamics modeling plays a positive role in improving prediction accuracy.

Further examining AGCRN and MTGNN, both models show more notable improvements in the R² metric, reaching 0.941 and 0.948 respectively. This suggests that they are better at modeling node heterogeneity and understanding global graph structures. In particular, MTGNN benefits from its multi-scale graph modeling strategy, which effectively captures the deeper structural patterns of traffic evolution and reduces prediction deviation.

By comparison, the proposed GCN+GRU model achieves the best results across all metrics. Specifically, MAE and RMSE are reduced to 2.01 and 4.12, while $R^2$ increases to 0.956. This indicates that the model demonstrates strong expressiveness in jointly capturing spatial features and modeling temporal sequences. It confirms that the effective integration of GCN and GRU can fully capture the spatiotemporal evolution patterns of traffic in complex topologies, thus enabling more accurate predictions.

Furthermore, this paper presents a comparative experiment between GRU and other time modeling methods, and the experimental results are shown in Figure 2.

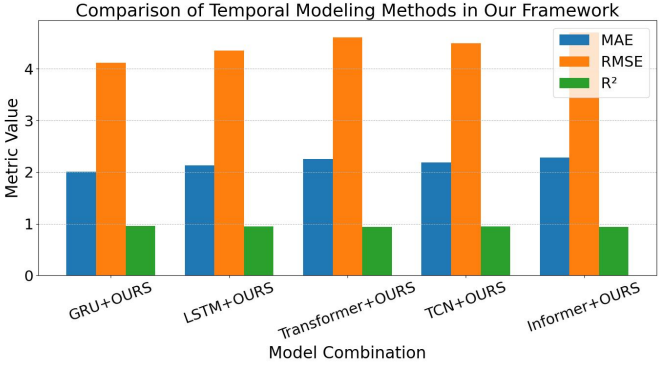

Figure 2. Comparative experiments on GRU and other time modeling methods

From the experimental results shown in the figure, it is evident that GRU differs significantly from other mainstream temporal modeling methods when combined with the graph convolution framework. The GRU+OURS model achieves the best performance in both MAE and RMSE, with the lowest values of 2.01 and 4.12, respectively. This indicates that it has stronger expressive power and generalization ability in temporal feature modeling, enabling more accurate capture of dynamic traffic patterns.

In contrast, methods such as LSTM, Transformer, and Informer exhibit some temporal modeling capabilities, but overall perform worse than GRU within this experimental framework. In particular, Transformer and Informer, despite their complex structures and ability to model long-range dependencies, show degraded performance on traffic data, which is characterized by strong local temporal patterns. Their MAE and RMSE are both higher than those of GRU, suggesting potential issues of overfitting or optimization difficulty in such tasks.

In terms of the $R^2$ metric, GRU+OURS also reaches the highest value of 0.956, further validating its superiority in overall fitting quality. In summary, GRU not only offers a simple structure and stable convergence within the proposed graph convolution prediction framework, but also outperforms current mainstream deep temporal modeling methods in practical prediction accuracy. It is well-suited to serve as the core temporal feature extraction module in complex network scenarios.

Finally, this paper gives a comparison of prediction effects under different adjacency matrix construction methods, and the experimental results are shown in Figure 3.

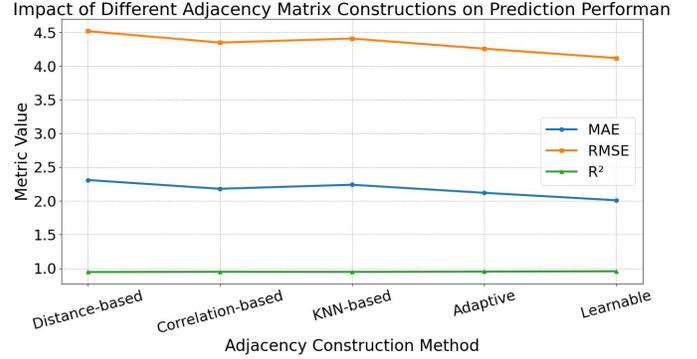

Figure 3. Comparison of prediction effects under different adjacency matrix construction methods

As shown in the figure, different adjacency matrix construction methods have a significant impact on network traffic prediction performance. Traditional distance-based and correlation-based methods perform relatively poorly in terms of MAE and RMSE, reaching 2.31/4.52 and 2.18/4.35, respectively. This indicates their limitations in capturing the true dependency relationships among nodes. Although these static adjacency methods are easy to implement, they fail to reflect the time-varying correlations between nodes, resulting in limited improvements in overall model performance.

With enhanced adjacency matrix construction methods, the performance of KNN-based and adaptive approaches shows gradual improvement. In particular, the adaptive method, which dynamically updates adjacency relationships, achieves better results in the $R^2$ metric, reaching 0.951. This suggests that introducing a structure-adaptive mechanism can improve the model's ability to represent heterogeneous topologies, thus enabling more effective spatial modeling and enhancing prediction accuracy.

The best results are obtained with the learnable adjacency method, which achieves MAE, RMSE, and $R^2$ values of 2.01, 4.12, and 0.956, respectively. This method allows the model to automatically learn the optimal adjacency structure during training, avoiding biases introduced by manual design. It ensures a tight coupling between spatial structure and prediction tasks. The results demonstrate that constructing a learnable adjacency matrix is an effective way to improve the performance of graph neural network-based prediction in complex network environments.

IV. CONCLUSION

This study presents a combined modeling approach utilizing Graph Convolutional Networks and Gated Recurrent Units to predict network traffic within complex topological structures. The method fully integrates spatial dependencies within the network structure and temporal dynamics in traffic sequences. It captures the information transmission process between nodes more comprehensively, enabling more accurate traffic prediction. The experimental findings indicate that the proposed model achieves superior performance compared to mainstream approaches on various evaluation metrics, confirming its effectiveness in managing complex and dynamic environments. In the modeling process, systematic comparative experiments are conducted on the number of graph convolution layers, the choice of temporal modeling methods, and strategies for constructing adjacency matrices. Results indicate that reasonable adjacency matrix construction and careful design of temporal modules have a significant impact on prediction performance. Among them, the model using a learnable adjacency structure achieves the best overall performance, suggesting that adaptive structural modeling provides clear advantages in heterogeneous and dynamic network environments.

Additionally, ablation studies and comparative experiments under various settings are conducted to further verify the generalization and robustness of the model. In combined tests with LSTM, Transformer, and Informer, GRU demonstrates better stability and faster convergence, proving its practicality in traffic prediction tasks. The proposed method is not only applicable to communication network traffic modeling but can also be extended to other graph-structured time series prediction tasks, such as intelligent transportation and power load forecasting. Future research can be extended in two directions. The model can be extended into a multi-modal input framework by incorporating additional features such as topology variations and transmission delays to enhance prediction robustness. Additionally, integrating more expressive large-scale pre-trained models and federated learning mechanisms can support broader modeling capabilities and enable secure collaborative prediction across heterogeneous cross-domain networks.